\documentclass{article}

\usepackage{microtype}
\usepackage{graphicx}
\usepackage{subfig}
\usepackage{booktabs} 
\usepackage[nounderscore]{syntax}
\newcommand{\paramspace}{\mathcal{S}}
\usepackage{amssymb}
\usepackage{amsmath} 
\usepackage{tabularx}
\newcolumntype{C}{>{\centering\arraybackslash}X}

\usepackage{hyperref}
\usepackage{cleveref}
\usepackage{xargs}
\usepackage[colorinlistoftodos,prependcaption]{todonotes}
\usepackage{makecell}
\newcommandx{\sg}[2][1=]{\todo[inline,#1]{{\normalfont \bfseries{SG}}: #2}}

\usepackage[accepted]{icml2021}

\icmltitlerunning{Inferring the Structure of Ordinary Differential Equations}

\begin{document}

\twocolumn[
\icmltitle{Inferring the Structure of \\
           Ordinary Differential Equations}



\icmlsetsymbol{equal}{*}

\begin{icmlauthorlist}
\icmlauthor{Juliane Weilbach}{bcai}
\icmlauthor{Sebastian Gerwinn}{bcai}
\icmlauthor{Christian Weilbach}{ubc}
\icmlauthor{Melih Kandemir}{bcai}
\end{icmlauthorlist}

\icmlaffiliation{ubc}{University of British Columbia, Vancouver, Canada}
\icmlaffiliation{bcai}{Bosch Center for Artificial Intelligence, Renningen, Germany}

\icmlcorrespondingauthor{Juliane Weilbach}{Juliane.Weilbach@de.bosch.com}

\icmlkeywords{Machine Learning, Ordinary Differential Equations, Symbolic Regression, Structure Search}

\vskip 0.3in
]



\printAffiliationsAndNotice{\icmlEqualContribution} 

\begin{abstract}
Understanding physical phenomena oftentimes means understanding the underlying dynamical system that governs observational measurements. While accurate prediction can be achieved with black box systems, they often lack interpretability and are less amenable for further expert investigation.
Alternatively, the dynamics can be analysed via symbolic regression.
In this paper, we extend the approach by \cite{udrescu2020ai} called AI Feynman to the dynamic setting to perform symbolic regression on ODE systems based on observations from the resulting trajectories.
We compare this extension to state-of-the-art approaches for symbolic regression empirically on several dynamical systems for which the ground truth equations of increasing complexity are available. Although the proposed approach performs best on this benchmark, we observed difficulties of all the compared symbolic regression approaches on more complex systems, such as Cart-Pole.


\end{abstract}

\section{Introduction}
\label{submission}

In multiple disciplines ranging from climate analysis over epidemiology up to financial portfolio optimization, and forecasting, the interpretability of the inferred dynamical model has as crucial importance as its predictive accuracy. While black-box machine learning methods are capable of accurate forecasts \cite{frigola_gpssm2014,hegde19a}, effective methods to find interpretable solutions remain as an open research question.
In contrast to black-box regression, symbolic regression aims at explaining the dynamics of a target system by combinations of a self-explanatory set of basis functions. It is common to use genetic algorithms in both scientific community \cite{schmidt_distilling_2009,bernardino2011inferring,nicolau_learning_2014,Quade_2016} as well as commercial implementations\footnote{https://www.nutonian.com/products/eureqa} of symbolic regression.

Genetic algorithms are capable of uncovering relationships in complex search spaces. However, these methods tend to generate highly complex and hardly interpretable solutions at high computational cost as the complexity of the involved expressions are typically not considered. Current approaches employ neural networks to search for matching expressions \cite{martius2016extrapolation,sahoo2018learning,kim2019integration} in order to address these weaknesses.
Inspired by a human expert, a promising symbolic regression approach for static data from  \citet{udrescu_ai_2020} checks whether various physical properties are present in the data. It includes information about physical equations in the search, such as checking the physical units first or finding symmetry properties through a neural network. The algorithm, called the {\it AI Feynman} explores complex structures in the data by performing a structured combinatorial search in the space of symbolic expressions. Their subsequent work \cite{udrescu2020ai} improves the robustness of the solutions against noise by calculating Pareto-fronts. However, their approach refers not directly to dynamical systems but to general physical equations. In contrast, the method of Brunton et al. \cite{brunton_supporting_nodate} successfully discovers equations from data collected from a nonlinear dynamical system by sparse identification. They exploit the fact that dynamical systems have only a few decisive terms, which makes them sparse in the function space spanned by a potential overcomplete set of basis-functions. However, a weak point of the procedure is that the search space must be defined beforehand using a fixed, additive function basis, which must include all valid terms. Therefore, strong prior knowledge of the system is necessary to define the basis.
Encoding prior knowledge over symbolic expressions and combining it with observations in a Bayesian fashion is used by \citet{jin_bayesian_2020} where a distribution over a symbolic tree is constructed, however, only applied to static, non-dynamic settings. 



In this paper, we propose an extension of \cite{udrescu2020ai} to the domain of dynamic data, thereby symbolically learning ODEs. In contrast of learning a static function, we use the method in \cite{udrescu2020ai} to find a symbolic expression of the right hand side of the ODE by providing sequences of states and finite-difference approximations similarly as in \cite{brunton2019data}. We compare the resulting approach against other state-of-the art symbolic regression techniques on a benchmark set of dynamical systems.


While all compared algorithms are able to identify simple systems such as the Lotka-Volterra system, the increasing difficulty within the ground truth system, e.g., Cart-Pole, is also reflected by the performance of the algorithms. Nevertheless, our approach manages to identify the dominant behaviour even in these examples, showing the potential of the presented approach.



\section{Problem Formulation} \label{sec:problem}
Given a sequence of observations $x_{t_0}, \dots x_{t_n}$ with $x_i \in \mathbb{R}^k$, we are interested in identifying the Ordinary Differential Equation (ODE) that describes the dynamical system from which the observations have been collected. More precisely, we aim at inferring  $f: \mathbb{R}^{k} \mapsto \mathbb{R}^k$ corresponding to the following initial value problem:
\begin{align}
\dot{x} &=  f(x(t)), \quad  x(t_{\text{\tiny{init}}}) = x_0, \label{eq:ode}
\end{align} 
where $x(t) \in \mathbb{R}^k$ is a dependent variable representing the state and $t$ is the time. In order to infer a description of the right hand side of the differential equation, we aim at finding a simple and interpretable characterization of the function $f$ in terms of basis functions. In contrast to black-box regression using non-parametric regression \cite{hegde19a,frigola_gpssm2014,ialongo2019overcoming}, we aim at a parametric functional expression.

We initially define a space of feasible symbolic expressions that can describe a dynamical system, in which we will perform the search. We define the search space in terms of a grammar that follows an explicit syntax. The following grammar characterizes the expressions in a search space $\paramspace\subset \{f: \mathbb{R}^k \rightarrow \mathbb{R}\}$ for a single dimension in the ODE:
\setlength{\grammarparsep}{20pt plus 1pt minus 1pt} 
\setlength{\grammarindent}{12em} 

\begin{grammar}\label{eq:grammar}
<expr> ::= <expr> <op> <expr> 
\alt <unit-op><expr>
\alt <var>

<op> ::= + | - | $\cdot$ | / | $\hat{\mkern6mu}$

<unit-op> ::= $\sin$ | $\cos$ | $\log$ |\dots
\alt $\exp$ 
\alt $identity$ 

<var> ::= $x_1$ | \dots | $x_k$ | c
\end{grammar}

Although such search space can be constructed for each of the dimensions separately,  we focus on a single dimension and assume the other dimensions to be fixed.

\section{Approach}
Our solution builds upon the AI Feynman method \cite{udrescu_ai_2020,udrescu2020ai}, which has been explored so far only in the context of fundamental physical laws and not for detailed system design for a dedicated application. Below, we provide a brief background on AI Feynman and refer the reader to  \cite{udrescu_ai_2020,udrescu2020ai} for details.

AI Feynman operates on data consisting of input data ${\bf{x}}_i = (x_1,\dots, x_k)_i, i=1,\dots,N$ and the corresponding labels $y_i$ obtained by $y_i=f({\bf{x}}_i)$ via an unknown mapping $f$. The algorithm finds a symbolic expression by iteratively breaking down the search of functions spanned by a grammar as mentioned in section~\ref{sec:problem} into smaller problems consisting of functions acting only on a subset of variables.

Additionally to the fitting error of the found solution on the available training data, AI Feynman also computes a complexity description length, reflecting the number of functions and operators in the found expression. Across those two losses, a Pareto-frontier is computed. To compare against other symbolic regression techniques (see section~\ref{sec:experiments}), we chose the found solution with the lowest fitting error.

Our key observation is that AI Feynman has not yet been used to identify time-dependent differential equations within a dynamical system. To extend the approach to such cases, we used a similar strategy as \citet{brunton2019data}. By  providing finite differences $\frac{{\bf{x}}_i(t) - {\bf{x}}_i(t-\Delta t)}{\Delta t}$ and corresponding inputs ${\bf{x}}(t)$ instead of measurements of the time-derivative, we used AI-Feynman to find a symbolic expression for the right hand side of the ODE. Therefore, the mismatch between time-derivative and finite difference is treated as measurement noise. In the following experiments, we refer to this dynamic extension as {\it DynAIFeynman}.

\section{Experiments}\label{sec:experiments}
\begin{figure*}[h!]
    \centering
    \subfloat[Lotka Volterra]{\label{fig:lotka}
    \includegraphics[width=0.45\linewidth]{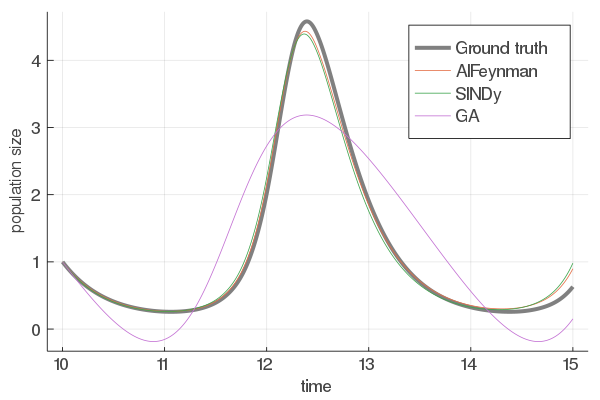}}
%
\subfloat[Cart-Pole]{\label{fig:cartpole}
    \includegraphics[width=0.45\linewidth]{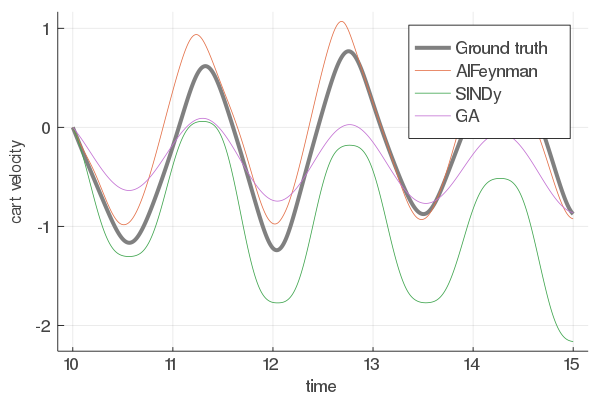}}
    \caption{Comparison of ground truth dynamics and inferred dynamics for different considered algorithms.}
    \label{fig:lodtka-voltera-double}
\end{figure*}

\begin{table*}[t]
\caption{RMSE scores for different methods evaluated on benchmark systems.}
\label{tab:results}
\vskip 0.15in
\begin{center}
\begin{small}
\begin{sc}
\begin{tabular}{lcccc}
\toprule
Method & Lotka-Volterra & Simple Pendulum & Cart-Pole\\
\midrule
DynAIFeynman (ours) & 0.19 $\pm$ 0.05 & 2.41 $\pm$ 0.13 & 1.23 $\pm$ 0.36 &  \\
GA-Baseline (ours)  & 2.13$\pm$ 1.11 & 2.47 $\pm$ 0.003 & 0.771 $\pm$ 0.79 &        \\
SINDy \cite{kaheman2020sindypi} & 0.24 & 2.19 & 1.92 &\\
\bottomrule
\end{tabular}
\end{sc}
\end{small}
\end{center}
\vskip -0.1in
\end{table*}

We compared the performance of our DynAIFeynman to other state-of-the-art symbolic regression methods on a benchmark of dynamical systems for which ground truth is available:

 \paragraph{Genetic algorithm}
Current state-of-the art \cite{schmidt_distilling_2009,Schmidt2010} in symbolic regression is often based on genetic algorithms which search the increasingly complex search space spanned by a grammar over functions (see Section~\ref{sec:problem}). Genetic algorithms are also used in commercially available software for symbolic regression \cite{eureqa}. To compare our proposed method against symbolic regression based on genetic algorithms, we implemented such a baseline using the following genetic algorithm to search the space of the grammar~\ref{eq:grammar}.
A function corresponding to an admissible word within the grammar~\ref{eq:grammar} is represented by a bitstring similar to \cite{bernardino2011inferring}. The genetic algorithm explores this  space of finite-length bitstrings to find a symbolic equation of the ODE which generated the available data.
Initially, a population of $N$ individuals is initialized  random bitstrings (each bit is set to 1 with probability 0.5), where $N$ is set individually per experiment to account for the different complexities of the dynamical systems (see Appendix for details).  




The bitstring encodes a sequence of grammar-rules to be applied in order to arrive at a symbolic experssion. As we used a finite bitstring size, it could happen that a sequence of rules does not lead to a valid symbolic expression as terminal expressions are still missing, e.g., a variable as argument of an earlier unit-op. When this happens during mutation or initialization of the population new initialization or mutation is generated until a valid expression is generated.The bitstring length is a hyperparameter and requires prior knowledge on how many symbols are necessary to generate the equation. 

Within one iterations, we first calculate, the fitness function, the RMSE for every candidate $f$ within the population:
\begin{align*}
    \mathcal{L}({\bf{x}}(t_i)) & = \sqrt{\frac{1}{N}\sum_i \left(f({\bf{x}}(t_{i})) - \frac{({\bf{x}}(t_{i+1}) - {\bf{x}}(t_i))}{(t_{i+1}-t_i)}\right)^2}
\end{align*}
In order to make the next generation more successful than the previous one, the best 50\% candidates (according to the above loss) were selected and transferred to the next generation. 
We ensured diversity from one generation to the next by applying mutation on the best individuals. We added these mutated individuals also to the next generation. Within the mutation procedure each bit in the string is flipped with a probability of 10\% leading to a new individual.

Despite providing a consistent feasible set for possible expressions, the grammar limits the search space due to the finite bitstring used. To allow for arbitrary coefficients in polynomials or frequencies in cosines, we could use regression coefficients in the grammar which we fit in a subsequent regression step. For the sake of simplicity, however, we modified the grammar such that it directly contains  the functions with coefficients from the ground truth of the experiments to avoid the additional regression step. In the experiments, we refer to this implementation as GA-Baseline. Note that the modification of the grammar results in an optimistic estimate of the baseline's performance, as the ground truth coefficients are not necessarily exactly known in practise.

\paragraph{SINDy}

In Sparse Identification of Nonlinear Dynamics (referred by SINDy in Table~\ref{tab:results}), a large matrix of $p$ basis functions $\Phi: (\mathbb{R}^k \rightarrow \mathbb{R})^p$ is constructed. We used the open source implementation of \cite{datadriven} for our experiments with SINDy. By means of linear superposition $\Phi {\bf{W}}$ with parameter ${\bf{W}} \in \mathbb{R}^{p \times k}$, a candidate function is constructed which can be evaluated on the given states ${\bf{x}}(t_i)$ representing the right hand side of the target ODE. As loss SINDy uses the LASSO regularized  mean squared error of the predictions obtained with such a weighted combination of basis functions.
 
  As only combinations of the basis functions can be inferred, prior knowledge is essential in designing a suitable set of functions representing the right hand side of the ODE. In particular, nested terms have to be explicitly encoded into the basis set $\Phi$ as nested functions cannot be generated by linear superposition.

\subsection{Dynamical systems}
We evaluated each of the competitors on the following dynamical systems of increasing dimensionality, thereby also reflecting increasingly complex behaviour.

\begin{description}
\item[Lotka Volterra]: This system consists of the following ODE:
\begin{align}
\label{eq:lv}
    \dot{x} & = x(\epsilon_1 -  y); &\epsilon_1 = 1.5, \\
    \dot{y} & = -y(\epsilon_2 -  x); &\epsilon_2 = 3
\end{align}
\item[Simple Pendulum]
\begin{align}
    \label{eq:sp}
    \dot{\theta}_1 & = \theta_2, \\
    \dot{\theta}_2 & = -\left(\frac{b}{m}\right)\theta_2 -\frac{g}{l} \sin(\theta_1)
\end{align}
\begin{align*}
    \text{with} \quad b=1,m=10,l=1,g=9.81.
\end{align*}
\item[Cart-Pole]
\begin{align}
\label{eq:cart-pole}
    \ddot{\theta} &= -\frac{(M + m)g \cdot \text{sin}(\theta)+ F\cdot l \cdot \text{cos}(\theta)}{l^{2}(M+m-m \cdot \text{cos}(\theta)^{2})} \nonumber \\
     \quad &+  \frac{m \cdot l^{2}\cdot \text{sin}(\theta)\cdot \text{cos}(\theta)\cdot\dot{\theta}^{2}}{l^{2}(M+m-m \cdot \text{cos}(\theta)^{2})} \\
    \ddot{x} &= \frac{m \cdot l^{2}\cdot\text{sin}(\theta) \cdot \ddot{\theta}^{2} + Fl}{l(M+m-m\cdot \text{cos}(\theta)^2)} \nonumber \\
     \quad &+ \frac{m \cdot g \cdot \text{sin}(\theta) \text{cos}(\theta)}{l(M+m-m\cdot \text{cos}(\theta)^2)} 
\end{align}

\end{description}
with $m=M=l=1$, $g=9.81$ and the control input as $F= - 0.2+0.5 \cdot sin(6t)$. To solve the Cart-Pole system including the optimal control, the kinematics $(x \quad \dot{x} \quad \theta \quad \dot{\theta})^T$ were derived by an \cite{kaheman_sindy-pi_2020}. 
As training data, we generated states with a time-discretization of $t_0=0, \dots, t_n = 10$ and provided finite state difference to the algorithms. 
To assess the quality of a symbolic estimate $\hat{f}$ of the right hand side  of \eqref{eq:ode}, we compute the mean squared error across states of the integrated ground truth solution $f$ using the discretization (see Appendix for details on the discretization scheme). 

As different dimensions in the ODE might have different scales, we only assessed the quality of the the last dimension for each dynamical system (these are also the ones shown in Figure~\ref{fig:lodtka-voltera-double})

Consequently, the test-error is computed by:
\begin{align}
\label{eq:test-error}
    \mathcal{E} & = \sqrt{\frac{1}{N}\sum_i \left(\hat{f}({\bf{x}}(\tau_i)) - f_k({\bf{x}}(\tau_i))\right)^2 }.
\end{align}

The evaluation of these competitor w.r.t. test-errors (according to \eqref{eq:test-error}) on these benchmark systems can be found in Table~\ref{tab:results}, whereas the estimated evolutions are plotted for selected dynamical systems in Figure~\ref{fig:lodtka-voltera-double}. As can be seen from Table~\ref{tab:results}, our proposed method is able to identify the fundamental behaviour of the underlying systems. 
Although none of the methods could identify relatively complex systems, such as the Cart-Pole system with high accuracy, our approach performs significantly better than SINDy and overlaps with the optimistic results from the GA-Baseline showing the potential of our proposed approach.


\section{Conclusion}

Symbolic regression offers an attractive alternative to black-box regression to gain insights into the mechanism driving a dynamical system by observing it. While black box regression can achieve accurate predictions, they need non-interpretable mathematical expressions such as neural networks for the explanation. In this paper, we explored symbolic regression in the context of dynamical systems. While simple systems can be accurately inferred by all competitors, we found out that complex systems consisting of multiple interacting variables could not be accurately identified, even when part of the ODE is provided as additional knowledge and no measurement noise is assumed on the observations. The only source of noise is the  time-discretized numerical integration. 
Our conjecture is that modelling the error by a black-box neural network in order to allow for non-Gaussianity of the error distribution could lead to promising hybrid modelling approaches. Investigation of this direction is left for future research.

\clearpage

\bibliography{main}
\bibliographystyle{icml2021}


\newpage
\onecolumn
\appendix

\section{Appendix}
In this appendix, we provide more details on the experimental setup we used for the different dynamical systems and also list the symbolic expressions found in the different repetitions (5 per dynamical system).

To illustrate the results for the Cart-Pole system, we use the following variables to reflect the transformation into a first order ODE:
\begin{align*}
    \theta & =w \\
    x &= x \\
    \dot{\theta} &= y \\
    \dot{x} &= z 
\end{align*}

\subsection{Generating training and test data for the different dynamical systems}
To generate data (training and test) we used the  \href{https://diffeq.sciml.ai/stable/solvers/ode_solve/#ode_solve}{Julia ODE Solver} with the following settings.

\subsubsection{Simple Pendulum}
Initial Values: $[0.4 \cdot  \pi; 1.0]$ \\
(Training) timespan: $(0.0, 10.0)$ \\
(Test) timespan: $(10.0, 15.0)$ \\
solver:  Tsitouras 5/4 Runge-Kutta method, default for non-stiff problems

\subsubsection{Lotka Volterra}
Initial Value: $[1.0,1.0]$ \\
(Training) timespan: $(0.0, 10.0)$ \\
(Test) timespan: $(10.0, 15.0)$ \\
solver: Tsitouras 5/4 Runge-Kutta method, default for non-stiff problems

\subsubsection{Cart-Pole}
Initial Values: $[0.3; 0; 1.0; 0]$ \\
(Training) timespan: $(0.0, 10.0)$ \\
(Test) timespan: $(10.0, 15.0)$ \\
solver:  Tsitouras 5/4 Runge-Kutta method, default for non-stiff problems

\subsection{Genetic Algorithm}

\begin{table}[!htbp]
\caption{Hyperparameter setting of the GA}
\label{tab:hyperparameter}
\begin{center}
\begin{small}
\begin{sc}
\begin{tabular}{lcccc}
\toprule
 & Lotka-Volterra & Simple Pendulum & Cart-Pole\\
\midrule
bitstring length  & 20 & 20  &  60  \\
\# of candidates  &  70 & 70  & 100   \\
iterations      & 100 & 40  &  100  \\
\bottomrule
\end{tabular}
\end{sc}
\end{small}
\end{center}
\end{table}
 \newpage
\subsubsection{Simple Pendulum Symbolic Expressions}
 
\begin{table}[!htbp]
\caption{Simple Pendulum results}
\label{tab:GA-SP-res}
\begin{center}
\begin{small}
\begin{sc}
\begin{tabular}{lcc}
\toprule
Runs & Symbolic & Test Loss \\
\midrule
1.  &  $\text{identity}(\theta_2 \cdot (-9.81)+(-0.1))$ & 2.476\\
2.  &  $(-1.0)+ \theta_2 \cdot (-9.81)$  & 2.471\\
3.  &  $(-1.0)+ \theta_2 \cdot (-9.81)$ & 2.471 \\
4.  &  $\text{identity}(\theta_2 + \theta_2 \cdot (-9.81))$ & 2.479  \\
5.  &  $-1.0)+\theta_2 \cdot (-9.81)$ & 2.471  \\
\bottomrule
\end{tabular}
\end{sc}
\end{small}
\end{center}
\end{table}
 
\subsubsection{Lotka Volterra Symbolic Expressions}
 
\begin{table}[!htbp]
\caption{Lotka Volterra results}
\label{tab:GA-LV-res}
\begin{center}
\begin{small}
\begin{sc}
\begin{tabular}{lcc}
\toprule
Runs & Symbolic & Test Loss \\
\midrule
1.  &  $\text{identity}(x+(-3.0)^{1.0})$ & 1.035\\
2.  &  $x+(-3.0)^{1.0}$  & 1.035\\
3.  &  $x+y\cdot (-3.0)$ & 2.484 \\
4.  &  $y^{cos(y+1.0)}$ & 3.632  \\
5.  &  $x+(-3.0)\cdot  y$ & 2.484  \\
\bottomrule
\end{tabular}
\end{sc}
\end{small}
\end{center}
\end{table}

\subsubsection{Cart Pole Symbolic Expressions}
 
\begin{table}[!htbp]
\caption{Cart Pole results}
\label{tab:GA-CP-res}
\begin{center}
\begin{small}
\begin{sc}
\begin{tabular}{lcc}
\toprule
Runs & Symbolic & Test Loss \\
\midrule
1.  &  $(-1.0)\cdot 6.0\cdot w/0.5$ & 2.168\\
2.  &  $(-1.0)\cdot w/0.5+z/y\cdot z/(-1.0)$  & 0.552\\
3.  &  $(-1.0)\cdot 2.0\cdot w$ & 0.417 \\
4.  &  $(-1.0)\cdot w/0.5^{\text{identity}(2.0)}$ & 0.301  \\
5.  &  $2.0\cdot (-1.0)\cdot w$ & 0.417 \\
\bottomrule
\end{tabular}
\end{sc}
\end{small}
\end{center}
\end{table}
\newpage
 
\subsection{DynAIFeynman results}
For DynAIFeynman for each system we used 500 training epochs for the interpolating neural network and polynomial fits are searched for up to degree 4. 
 
\subsubsection{Simple Pendulum Symbolic Expressions}
 
\begin{table}[hbt!]
\caption{Simple Pendulum results}
\label{tab:AI-SP-res}
\begin{center}
\begin{small}
\begin{sc}
\begin{tabularx}{\linewidth}{c *{2}{C}}
\toprule
Runs & Symbolic & Test Loss \\
\midrule
1.  &  $-0.0324473939912423\cdot \theta_1^4 - 0.00172720766565721\cdot \theta_1^3\cdot \theta_2 + 1.47344025442332\cdot \theta_1^3 - 0.00422370746938726\cdot \theta_1^2\cdot \theta_2^2 + 0.0224633714266906\cdot \theta_1^2\cdot \theta_2 + 0.058925455225393\cdot \theta_1^2 - 0.00152727919744772\cdot \theta_1\cdot \theta_2^3 - 0.00390131101287516\cdot \theta_1\cdot \theta_2^2 + 0.0065764122456723\cdot \theta_1\cdot \theta_2 - 9.74355034133818\cdot \theta_1 - 1.66932950476792e-5\cdot \theta_2^4 - 4.78902623248226e-5\cdot \theta_2^3 + 0.00455847612525189\cdot \theta_2^2 - 0.150903340299233\cdot \theta_2 - 0.0292019001025895$ & 2.167 \\
2.  &  $-0.00112081291984988\cdot \theta_1^3 - 0.0684746158057592\cdot \theta_1^2\cdot \theta_2 - 0.0178759339534681\cdot \theta_1^2 - 0.0772194176628979\cdot \theta_1\cdot \theta_2^2 + 1.76683800398161\cdot \theta_1\cdot \theta_2 + 0.0561062364127456\cdot \theta_1 - 0.00717025434174159\cdot \theta_2^3 - 0.0349584835773569\cdot \theta_2^2 - 3.28714749810262\cdot \theta_2 - 0.301521031753468$  & 2.473\\
3.  &  $ -0.0017513798882943\cdot \theta_1^3 - 0.0647354771913438\cdot \theta_1^2\cdot \theta_2 - 0.0116572879434979\cdot \theta_1^2 - 0.0744981362050606\cdot \theta_1\cdot \theta_2^2 + 1.72779223496299\cdot \theta_1\cdot \theta_2 + 0.0374990502052437\cdot \theta_1 - 0.00858538337680672\cdot \theta_2^3 - 0.0241147851789653\cdot \theta_2^2 - 3.27571089211806\cdot \theta_2 - 0.274026438170276$ & 2.472 \\
4.  &  $-0.00139154384443193\cdot \theta_1^3 - 0.065472337675063\cdot \theta_1^2\cdot \theta_2 - 0.0151266304275168\cdot \theta_1^2 - 0.0763910225267537\cdot \theta_1\cdot \theta_2^2 + 1.74376893724452\cdot \theta_1\cdot \theta_2 + 0.0456267320438428\cdot \theta_1 - 0.00725738312725596\cdot \theta_2^3 - 0.0281131065275104\cdot \theta_2^2 - 3.29247873163534\cdot \theta_2 - 0.279312281615113$ & 2.472  \\
5.  &  $-0.00138052368307217\cdot \theta_1^3 - 0.0672214030272411\cdot \theta_1^2\cdot \theta_2 - 0.0177039805578151\cdot \theta_1^2 - 0.077788931500498\cdot \theta_1\cdot \theta_2^2 + 1.75964827786662\cdot \theta_1\cdot \theta_2 + 0.0589767776019023\cdot \theta_1 - 0.00701982650483074\cdot \theta_2^3 - 0.0352782471195393\cdot \theta_2^2 - 3.27547564696099\cdot \theta_2 - 0.307167790003207$ & 2.473 \\
\bottomrule
\end{tabularx}
\end{sc}
\end{small}
\end{center}
\end{table}
 \newpage
 
\subsubsection{Lotka Volterra Symbolic Expressions}
 
\begin{table}[!htbp]
\caption{Lotka Volterra results}
\label{tab:AI-LV-res}
\begin{center}
\begin{small}
\begin{sc}
\begin{tabularx}{\linewidth}{c *{2}{C}}
\toprule
Runs & Symbolic & Test Loss \\
\midrule
1.  &  $-0.00112081291984988 \cdot x^3 - 0.0684746158057592 \cdot x^2 \cdot y - 0.0178759339534681 \cdot x^2 - 0.0772194176628979 \cdot x \cdot y^2 + 1.76683800398161 \cdot x \cdot y + 0.0561062364127456 \cdot x - 0.00717025434174159 \cdot y^3 - 0.0349584835773569 \cdot y^2 - 3.28714749810262 \cdot y - 0.301521031753468
$ & 0.258\\
2.  &  $-0.00152938837980895 \cdot x^3 - 0.0576300494028292 \cdot x^2 \cdot y - 0.0127993611378648  \cdot x^2 - 0.0666162927164024\cdot x \cdot y^2 + 1.65290648558145 \cdot x \cdot y + 0.047579473066381 \cdot x - 0.00763201564293955 \cdot y^3 - 0.02930836844511 \cdot y^2 - 3.22458355359839 \cdot y - 0.273077533676859
$  & 0.134\\
3.  &  $-0.0017513798882943 \cdot x^3 - 0.0647354771913438\cdot x^2 \cdot y - 0.0116572879434979\cdot x^2 - 0.0744981362050606\cdot x \cdot y^2 + 1.72779223496299 \cdot x \cdot y + 0.0374990502052437\cdot x - 0.00858538337680672\cdot y^3 - 0.0241147851789653 \cdot y^2 - 3.27571089211806\cdot y - 0.274026438170276
$ & 0.162 \\
4.  &  $-0.00139154384443193 \cdot x^3 - 0.065472337675063 \cdot x^2 \cdot y - 0.0151266304275168 \cdot x^2 - 0.0763910225267537 \cdot x \cdot y^2 + 1.74376893724452 \cdot x \cdot y + 0.0456267320438428 \cdot x - 0.00725738312725596 \cdot y^3 - 0.0281131065275104 \cdot y^2 - 3.29247873163534 \cdot y - 0.279312281615113
$ & 0.212  \\
5.  &  $-0.00138052368307217 \cdot x^3 - 0.0672214030272411 \cdot x^2 \cdot y - 0.0177039805578151 \cdot x^2 - 0.077788931500498 \cdot x \cdot y^2 + 1.75964827786662 \cdot x \cdot y + 0.0589767776019023 \cdot x - 0.00701982650483074 \cdot y^3 - 0.0352782471195393 \cdot y^2 - 3.27547564696099 \cdot y - 0.307167790003207
$ & 0.213  \\
\bottomrule
\end{tabularx}
\end{sc}
\end{small}
\end{center}
\end{table}

\newpage
 
\subsubsection{Cart Pole Symbolic Expressions}
 
\begin{table}[!htbp]
\caption{Cart Pole results}
\label{tab:AI-CP-res}
\begin{center}
\begin{small}
\begin{sc}
\begin{tabularx}{\linewidth}{c *{2}{C}}
\toprule
Runs & Symbolic & Test Loss \\
\midrule
1.  &  $-11.845962128067\cdot (w\cdot exp((cos(cos(y))-1)))$ & 1.386\\
2.  &  $4.359933909335\cdot (w\cdot (-exp(cos(cos(y)))))$  & 1.387\\
3.  &  $-4.360217092882\cdot (w\cdot exp(cos(cos(y))))$ & 1.387 \\
4.  &  $4.354924528126\cdot (w\cdot (-exp(cos(cos(y)))))$ & 1.384  \\
5.  &  $47.7689497990435\cdot w^4 + 34.8000295056234\cdot w^3\cdot x - 2.96481379197869\cdot w^3\cdot y - 10.0549064794547\cdot w^3\cdot z + 6.55594893070674\cdot w^3 - 3.27769585993784\cdot w^2\cdot x^2 + 21.6433852884368\cdot w^2\cdot x\cdot y - 46.9514382484925\cdot w^2\cdot x\cdot z + 16.8964156972663\cdot w^2\cdot x - 6.93282001830109\cdot w^2\cdot y^2 + 54.9368263810748\cdot w^2\cdot y\cdot z + 21.9768534384255\cdot w^2\cdot y - 69.4580871163974\cdot w^2\cdot z^2 - 46.9708759837534\cdot w^2\cdot z + 23.6865747302531\cdot w^2 - 4.64988536666163\cdot w\cdot x^3 - 36.3891291536006\cdot w\cdot x^2\cdot y + 36.7277843174582\cdot w\cdot x^2\cdot z - 4.86606558861306\cdot w\cdot x^2 + 6.89303461269625\cdot w\cdot x\cdot y^2 - 7.51384324101677\cdot w\cdot x\cdot y\cdot z - 70.30858960581\cdot w\cdot x\cdot y - 2.5732998566741\cdot w\cdot x\cdot z^2 + 71.2772539468098\cdot w\cdot x\cdot z + 1.00497137562927\cdot w\cdot x + 1.16678730685183\cdot w\cdot y^3 - 7.51763943025243\cdot w\cdot y^2\cdot z + 3.16958301802692\cdot w\cdot y^2 + 13.6174710214547\cdot w\cdot y\cdot z^2 - 6.05300362527973\cdot w\cdot y\cdot z - 32.3362512073897\cdot w\cdot y - 9.07347204852936\cdot w\cdot z^3 - 2.91912427675261\cdot w\cdot z^2 + 35.2955753892911\cdot w\cdot z - 5.18058514451248\cdot w - 6.13493811289436\cdot x^4 + 22.23915768717\cdot x^3\cdot y - 19.0614115496817\cdot x^3\cdot z - 34.6130891815334\cdot x^3 - 2.22277748363875\cdot x^2\cdot y^2 - 7.43145345365155\cdot x^2\cdot y\cdot z + 70.5942236381929\cdot x^2\cdot y + 17.7387079764248\cdot x^2\cdot z^2 - 69.5439160953327\cdot x^2\cdot z - 66.0065984707298\cdot x^2 - 4.08039415665966\cdot x\cdot y^3 + 27.1381442663131\cdot x\cdot y^2\cdot z - 1.75402667647653\cdot x\cdot y^2 - 65.5115686701007\cdot x\cdot y\cdot z^2 - 20.7589578553513\cdot x\cdot y\cdot z + 74.0506404963519\cdot x\cdot y + 49.8068600921779\cdot x\cdot z^3 + 40.6767617298183\cdot x\cdot z^2 - 85.1464172277568\cdot x\cdot z - 52.0768152973954\cdot x - 1.93110829839297\cdot y^4 + 14.3785209042429\cdot y^3\cdot z - 6.53915016242934\cdot y^3 - 38.3097512010258\cdot y^2\cdot z^2 + 41.6062540813935\cdot y^2\cdot z - 0.0534975086740148\cdot y^2 + 45.5794418053017\cdot y\cdot z^3 - 94.2848380972032\cdot y\cdot z^2 - 10.0666250752058\cdot y\cdot z + 26.4167495963713\cdot y - 21.5410316614469\cdot z^4 + 68.5262250113664\cdot z^3 + 19.9095141422248\cdot z^2 - 35.8785826203587\cdot z - 14.1024059459915
$ & 0.579 \\
\bottomrule
\end{tabularx}
\end{sc}
\end{small}
\end{center}
\end{table}
\newpage
 
\subsection{SINDy}
To solve the optimization scheme within SINDy, we used the following optimizer \href{https://datadriven.sciml.ai/stable/sparse_identification/optimizers/}{DataDrivenDiffEq.jl}.

\subsubsection{Symbolic Expressions}
 
\begin{table}[!htbp]
\caption{SINDy results}
\label{tab:Sindy-res}
\begin{center}
\begin{small}
\begin{sc}
\begin{tabularx}{\linewidth}{c  *{5}{C}}
\toprule
System & Symbolic & Test Loss & optimizer & max. iterations \\
\midrule
Simple Pendulum  &  $ -1 \cdot (sin(\theta_1) \cdot 1.0 + -0.007 \cdot \theta_1 + 0.014 \cdot \theta_2) \cdot inv(0.100) ^1$  & 2.186 & SR3 & 5000\\
Lotka Volterra  &  $-1 \cdot (-1.0 \cdot x + -0.073 \cdot y + 0.666 \cdot (1.5x + -1 \cdot x \cdot y) + 0.691 \cdot x \cdot y) \cdot inv(-7.011e-5 \cdot x \cdot y + -0.023) ^ 1$ & 0.239 & ADMM & 10000\\
Cart Pole  &  $(-1 \cdot ((sin(w) \cdot 1.0 + -0.6940608327497251w) + (cos(w) \cdot sin(w)) \cdot -0.30461453860504384)) \cdot 1642.312493997001$ & 1.920 & SR3 & 5000\\
\bottomrule
\end{tabularx}
\end{sc}
\end{small}
\end{center}
\end{table}

\subsubsection{Basis for SINDy}
Each function $f$ represents one equation in the basis matrix $\Phi$.\\
\\
\bfseries{Simple Pendulum}
\begin{align*}
    \begin{split}
        f_1 = \theta_1 \\
f_2 = \theta_2 \\
f_3 = sin(\theta_1) \\
f_4 = cos(\theta_2) \\
f_5 = cos(\theta_1) \\
f_6 = sin(\theta_2) \\
f_7 = cos(\theta_1) \cdot sin(\theta_2) \\
f_8 = 1.0 
    \end{split}
\end{align*}
\bfseries{Lotka Volterra}
\begin{align*}
    \begin{split}
        f_1 = x \\
f_2 = y \\
f_3 = x \cdot y \\
f_4 = sin(x) \\
f_5 = sin(y) \\
f_6 = cos(x) \\
f_7 = cos(y) \\
f_8 = 1.0 \\
f_9 = 1.5x + -1 \cdot x \cdot y
    \end{split}
\end{align*}

\bfseries{Cart-Pole}
\begin{align*}
    \begin{split}
    f_1 = 1.0 \\
    f_2 = w \\
f_3 = x \\
f_4 = y \\
f_5 = z \\
f_6 = y ^ 2 \\
f_7 = z ^ 2 \\
f_8 = y ^ 3 \\
f_9 = z ^ 3 \\
f_{10} = y ^ 4 \\
f_{11} = z ^ 4 \\
f_{12} = sin(w) \\
f_{13} = cos(w) \\
f_{14} = sin(w) \cdot y \\
f_{15} = sin(w) \cdot z \\
f_{16} = sin(w) \cdot y ^ 2 \\
f_{17} = sin(w) \cdot z ^ 2 \\
f_{18} = cos(w) ^ 2 \\
f_{19} = cos(w) \cdot sin(w) \\
f_{20} = cos(w) \cdot sin(w) \cdot y \\
f_{21} = cos(w) \cdot sin(w) \cdot z \\
f_{22} = cos(w) \cdot sin(w) \cdot y ^ 2 \\
f_{23} = cos(w) \cdot sin(w) \cdot z ^ 2 \\
f_{24} = -0.2 + 0.5 \cdot sin(6.0t) \\
f_{25} = cos(w) \cdot (-0.2 + 0.5 \cdot sin(6.0t)) \\
f_{26} = sin(w) \cdot (-0.2 + 0.5 \cdot sin(6.0t)) \\
f_{27} = -1 \cdot (cos(w) \cdot (-0.2 + 0.5 \cdot sin(6t)) + 19.62 \cdot sin(w) + cos(w) \cdot sin(w) \cdot y ^ 2) \cdot inv(2 + -1 \cdot cos(w) ^ 2) ^ 1
    \end{split}
\end{align*}


%




\end{document}